\documentclass[10pt,twocolumn,journal]{IEEEtran}
\usepackage[]{fontenc}
\usepackage[latin9]{inputenc}
\usepackage{color}
\usepackage{amsmath}
\usepackage{amssymb}
\usepackage{graphicx}

\makeatletter
\usepackage[T1]{fontenc}
\usepackage{graphics} 
\usepackage{epsfig} 
\usepackage{epstopdf}
\usepackage{mathptmx} 
\usepackage{times} 
\usepackage{amsmath} 
\usepackage{mathtools}
\usepackage{amssymb}  
\usepackage{txfonts}
\usepackage{algorithm}
\usepackage{algorithmic}
\usepackage{tabu}
\usepackage[english]{babel}

\makeatother

\begin{document}

\title{Joint Communication-Motion Planning in Wireless-Connected Robotic
Networks: Overview and Design Guidelines}

\author{Bo Zhang, Yunlong Wu, Xiaodong Yi, Xuejun Yang\\
 \thanks{B.Zhang, Y.Wu, X.Yi and X.Yang are with the State Key Laboratory of
High Performance Computing (HPCL), School of Computer, National University
of Defense Technology, Changsha, 410073, China. Email: \{zhangbo10,
ylwu1988, yixiaodong, xjyang\}@nudt.edu.cn}\vspace{0.2cm}
 }
\maketitle
\begin{abstract}
\noindent Recent years have witnessed the prosperity of robots and
in order to support consensus and cooperation for multi-robot system,
wireless communications and networking among robots and the infrastructure
have become indispensable. In this technical note, we first provide
an overview of the research contributions on communication-aware motion
planning (CAMP) in designing wireless-connected robotic networks (WCRNs),
where the degree-of-freedom (DoF) provided by motion and communication
capabilities embraced by the robots have not been fully exploited.
Therefore, we propose the framework of joint communication-motion
planning (JCMP) as well as the architecture for incorporating JCMP
in WCRNs. The proposed architecture is motivated by the observe-orient-decision-action
(OODA) model commonly adopted in robotic motion control and cognitive
radio. Then, we provide an overview of the orient module that quantify
the connectivity assessment. Afterwards, we highlight the JCMP module
and compare it with the conventional communication-planning, where
the necessity of the JCMP is validated via both theoretical analysis
and simulation results of an illustrative example. Finally, a series
of open problems are discussed, which picture the gap between the
state-of-the-art and a practical WCRN.\end{abstract}

\begin{IEEEkeywords}
Joint communication-motion planning, robotics networks, learning (artificial
intelligence), fading channels, dynamic programming\vspace{0.2cm}

\end{IEEEkeywords}

\section{Introduction of JCMP and Related Works}

\label{sec:Introduction}

Recent years have witnessed the evolution of robotics and great industrial/academic
efforts. As the robots interact with the physical and social environments,
considerable research contributions have been devoted to robotic sensing,
cognition, motion/path planning and control\cite{Kehoe2015}. A multi-robot
system aims at achieving challenging tasks or significantly improving
mission performance compared with a single robot, which demands consensus
and cooperation among robots \cite{Olfati-Saber2007}. Therefore,
maintaining the connectivity quality for information exchange among
robots becomes vital. As mobile robots are less likely to be connected
via wires, the wireless communications and networking among robots
and the infrastructure would play an crucial role and the wireless-connected
robotic networks (WCRNs) are very likely to be incorporated into the
next-generation communication networks.

\subsection{Overview of CAMP}

There is growing interest in incorporating autonomous robots into
wireless communication networks\cite{Lindhe2009,Tekdas2009,Ghaffarkhah2011,Ghaffarkhah2012,Daniel2011,Kim2012,Fink2013,Kudelski2014,Gil2015},
and most contributions focuses on designing communication-aware motion
planning (CAMP) for different applications. 

In \cite{Lindhe2009}, Lindhe and Johansson suggested exploiting multi-path
fading by motion plan in order to achieve a higher channel capacity,
while Gil et al. utilized the directional signal strength information
to design a simple positional controller by adapting to wireless signals
in real-world environments\cite{Gil2015}. In \cite{Tekdas2009},
Tekdas et al. adopted mobile robots for collecting data from wireless
sensor networks (WSNs) in order to extend the lifetime of the sensor
system by reducing the communication energy consumption of the sensing
nodes. Ghaffarkhah and Mostofi proposed to exploit the mobility of
mobile robots for improving the performance of wireless channel assessment
and target tracking in\cite{Ghaffarkhah2011}, as well as minimizing
the probability of target detection error for surveillance, while
guaranteeing connectivity constraints in \cite{Ghaffarkhah2012}.
Kudelski et al proposed that a group of robots may exploit mobility
to effectively and rapidly learn the link quality model in an unknown
environment\cite{Kudelski2014}. Daniel et al considered the sensing
task by multiple coordinated unmanned ariel vehicles (UAVs) , where
both the self-organizing mesh networking and channel-aware mobility
control contributes to a more timely and accurate information data
collection and fusion. Kim and Seo proposed a spatially secure group
communication problem, where the mobility of the UAVs were planned
to occupy a smaller space in order to improve the group security,
while preventing from becoming a over-dense group to avoid communication
congestion\cite{Kim2012}. Fink et al. proposed to combine adaptive
routing and motion control for maximizing the probability of having
a connected network\cite{Fink2012,Fink2013}. Considering a mobile
robot visiting several point-of-interests while communicating with
a base station, Yan and Ali et al aimed at minimizing the total energy
consumption and adopted mixed integer linear program \cite{Yan2014}
and the optimal control methods\cite{Ali2015}.

\begin{figure}
\centering{}\includegraphics[width=1\linewidth]{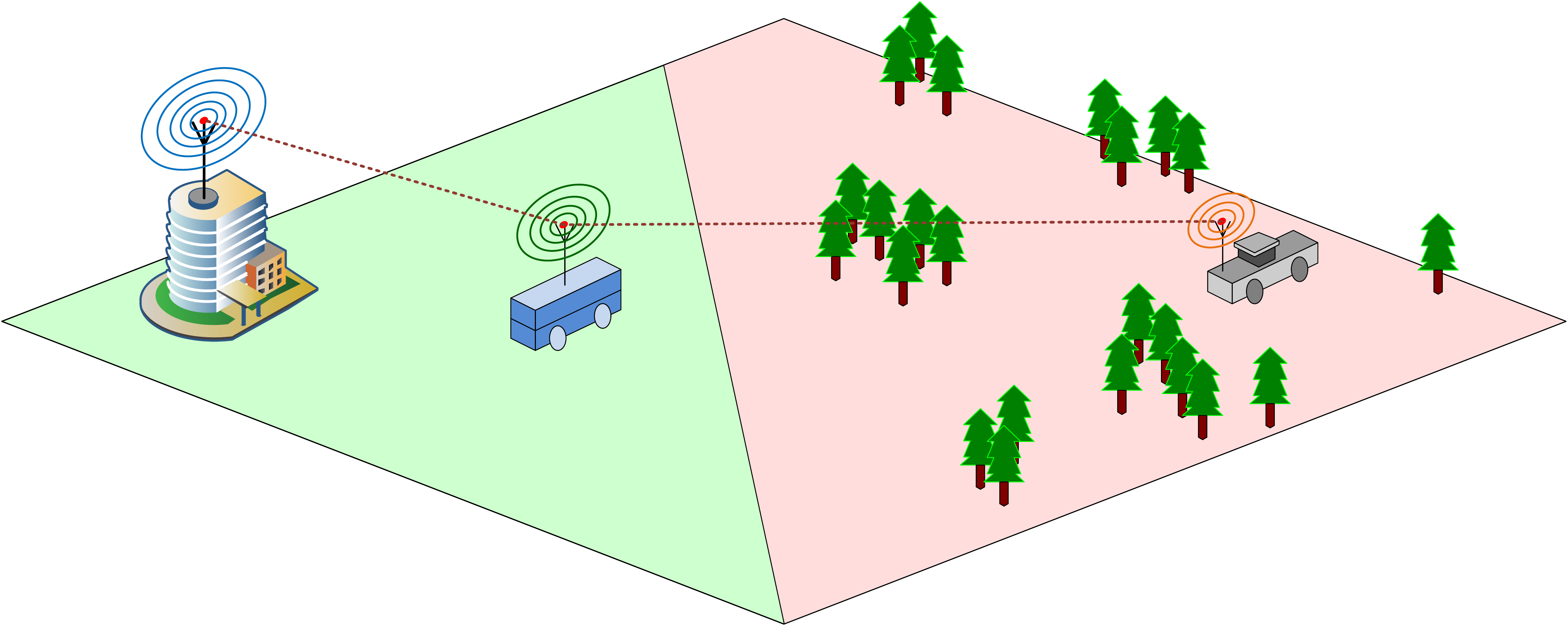} \protect\caption{The scenario of a sensing robot survey a sensitive area, and transmit
the monitored information to a base station.}
\label{fig:Example:System_Model}
\end{figure}

According to the above contributions, \textit{CAMP may be summarized
as to utilize the knowledge of connectivity quality for planning the
robotic motion in order to improve specified task-oriented performance,
while satisfying certain communication constraints}\cite{Ghaffarkhah2011}.
Therefore, CAMP focuses on exploiting the mobility resources in order
to optimize the motion plan, while the communication schemes are fixed
or having a limited adaptive capability so that the communication
quality has to be guaranteed by motion plan\cite{Yan2014}. Although
some works have considered adaptive transmission schemes such as adaptive
transmission rate\cite{Yan2014} and routing\cite{Fink2012}, \textit{they
are far away from fully exploiting the degree-of-freedom (DoF) in
space, time, frequency and energy dimensions in optimizing the communication
quality for supporting the mission objectives}. Let us take two simple
examples of robot surveillance in order to illustrate the above assertion.
The scenario is illustrated in Fig. \ref{fig:Example:System_Model},
in which a sensing robot explores the area and tracks any targets
of interest, then transmits the sensing data to the base station.

In the first example, the wireless channel quality is heavily degraded
due to the shadowing of an obstacle, e.g. several trees. In this scenario,
the sensing robot may exploit its mobile capability to move away from
the shadowing area and to find a better spot for communications\cite{Lindhe2009}.
Another solution is to deploy a router robot to build up a new relay-aided
link spanning from the sensing robot to the base station, which detours
the deep-shadowed communication channels\cite{Yan2012}. In this example,
the CAMP may greatly improve the communication quality by exploiting
the DoF in mobility. 

In the second example, the wireless channels between the sensing robot
and the base station is heavily interfered in the time-frequency domain.
In this case, exploiting the mobility of the robots or deploying router
robots may not achieve a noticeable improvement in the communication
quality. In contrast, self-adaptive sensing of the spectrum and changing
the communication frequency to a less-interfered channel may greatly
improve the communication quality. Another solution is to exploit
multiple antennas for beamforming a null point towards the interference
sources, which may also greatly improve the communication quality\cite{El-Hajjar2010}.
In this example, the CAMP is inferior to \textit{communication planning}
that exploits the frequency and spatial diversity.

\subsection{Evolution to JCMP}

Against this background, the evolution from CAMP to JCMP is inevitable,
where JCMP aims at\textbf{\textit{ joint exploiting the DoF in mobility,
space, time, frequency and energy (MSTFE) dimensions from both the
motion and communication components equipped by the robots}}. As a
result, the wireless-connected robotic networks adopting JCMP will
\textit{be capable of covering a wider range of application scenarios}
that involves interference and multi-access, etc.

After the differences between the CAMP and JCMP being identified,
several research questions are raised in order to be implement CAMP/JCMP
in a wireless-connected robotic networks (WCRNs), where the most imperative
ones are given as follows:
\begin{itemize}
\item How to develop an architecture for WCRN in which the JCMP may be incorporated?
\item How to quantify the connectivity quality in WCRNs?
\item How to effectively design the JCMP in order to exploit the DoF in
MSTFE dimensions?
\end{itemize}
The rest of the paper would be devoted to the above research questions.
Although we try to address the above research questions, it should
be noted that it is difficult to address the above questions in a
single technical note. Therefore, a trade-off has been made between
the proposed architecture design and necessary literature survey due
to the space limitations. 

In Section \ref{sec:OODA}, an OODA-based architecture is proposed
in order to address the first question, while in Section \ref{sec:connectivity_metric_assess},
the second question is approached by a brief overview of the research
on connectivity quality assessment in WCRNs. In \ref{sec:JCMP}, two
categories of JCMP designs are summarized and an illustrative design
example is given. Besides, several open problems are put forward along
with the conclusions in Section \ref{sec:future_works}.

\section{OODA-based WCRN Architecture}

\label{sec:OODA}

In this section, the architecture of the wireless-connected robotic
network is proposed, which is motivated by the observe-orient-decide-act
(OODA) cognitive-behavioral model\cite{Boyd2007} and its application
in robotic motion control and cognitive radio (CR). 

The observe-orient-decide-act (OODA) loop was first proposed by John
Boyd in the mid-1950s \cite{Boyd2007,Bryant2006}. Though Boyd initially
applied the concept to the military combat operations process, it
is now also often applied to a wide variety of areas, such as understanding
the commercial operations, the learning processes, etc\cite{Osinga2007}. 

A robotic motion control system may be decomposed into a series of
functional units, namely, the perception, modeling, planning, task
execution and motor control, where the perception is implemented by
the sensors for observing the environments, while the task execution
and motor control are carried out by the actuators to interact with
the environments\cite{Brooks1986}. Therefore, the robotic motion
control system may be modeled as a OODA loop. 

\begin{figure}[tb]
\centering{}\includegraphics[width=1\linewidth]{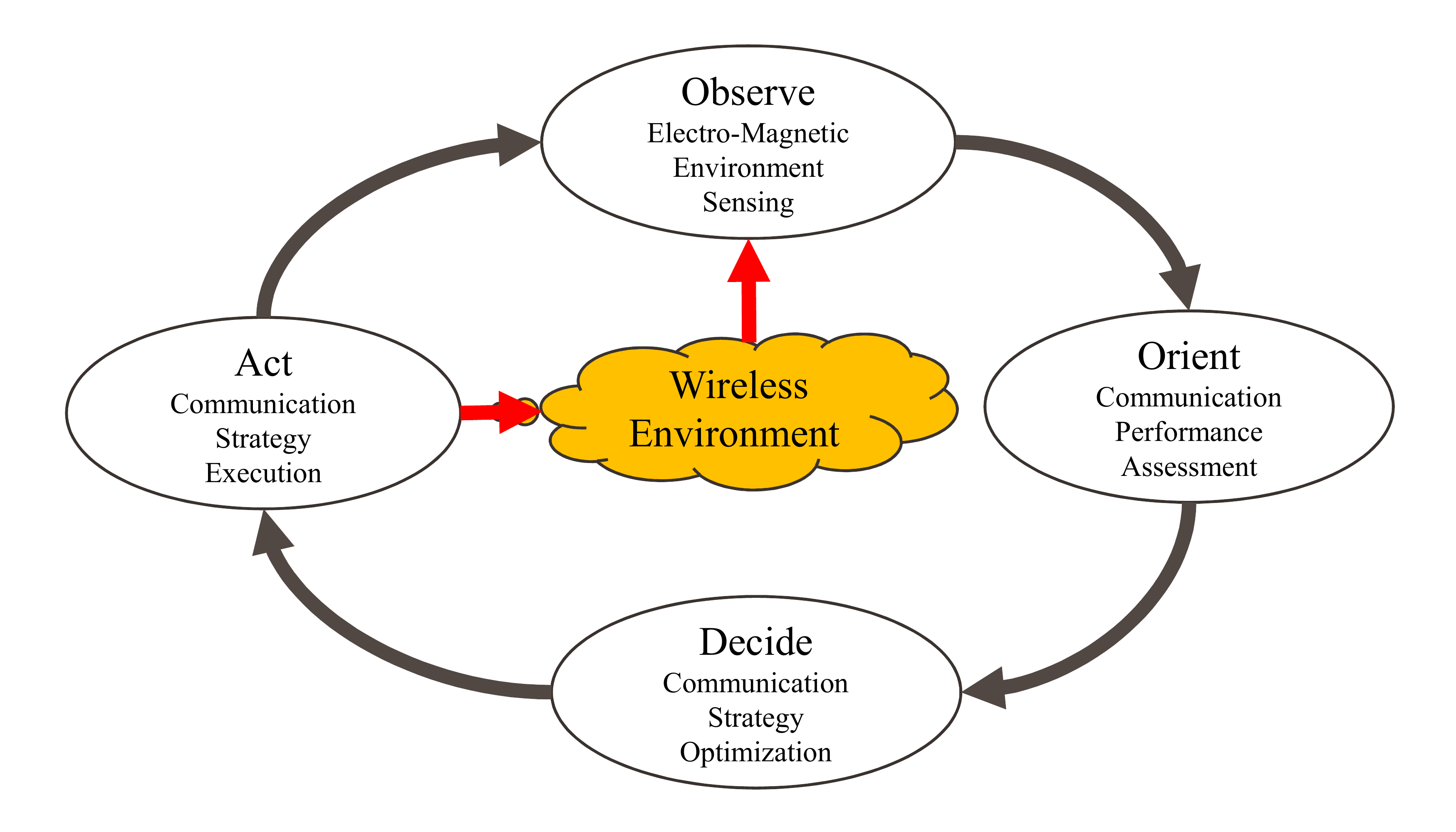} \protect\caption{OODA model for a cognitive radio system.}
\label{fig:OODA-CR}
\end{figure}

The adoption of OODA model in the area of wireless communications
may be traced back to 1999, when Mitola. etc propose the concept of
CR and explains the cognition cycle\cite{Mitola1999}. Jayaweera and
Christos proposed the concept of \textit{RadioBot} and treated the
radio device an equivalence of a robot in mechanical engineering\cite{Jayaweera2011}.
In the technical note summarizing the developments of CR in 2013\cite{Fette2013},
Fette explicitly related the CR cognition cycle to the OODA loop model,
including \textit{Observe }by measuring the elector-magnetic environment,
\textit{Orient} to the mission objective by adapting at different
protocol layers, \textit{Decide} by making good performance choices
for the mission, and \textit{Act} on the decisions, which is illustrated
in Fig. \ref{fig:OODA-CR}. It should also be noted that besides the
explicit observe, orient, decide and act procedures, the learning
capability is also highlighted in the CR, indicating a CR is capable
of continuously self-learning from its past experience in order to
improve the performance during the orient and the decide procedures.

\begin{figure*}[!t]
\centering{}\includegraphics[width=1\linewidth]{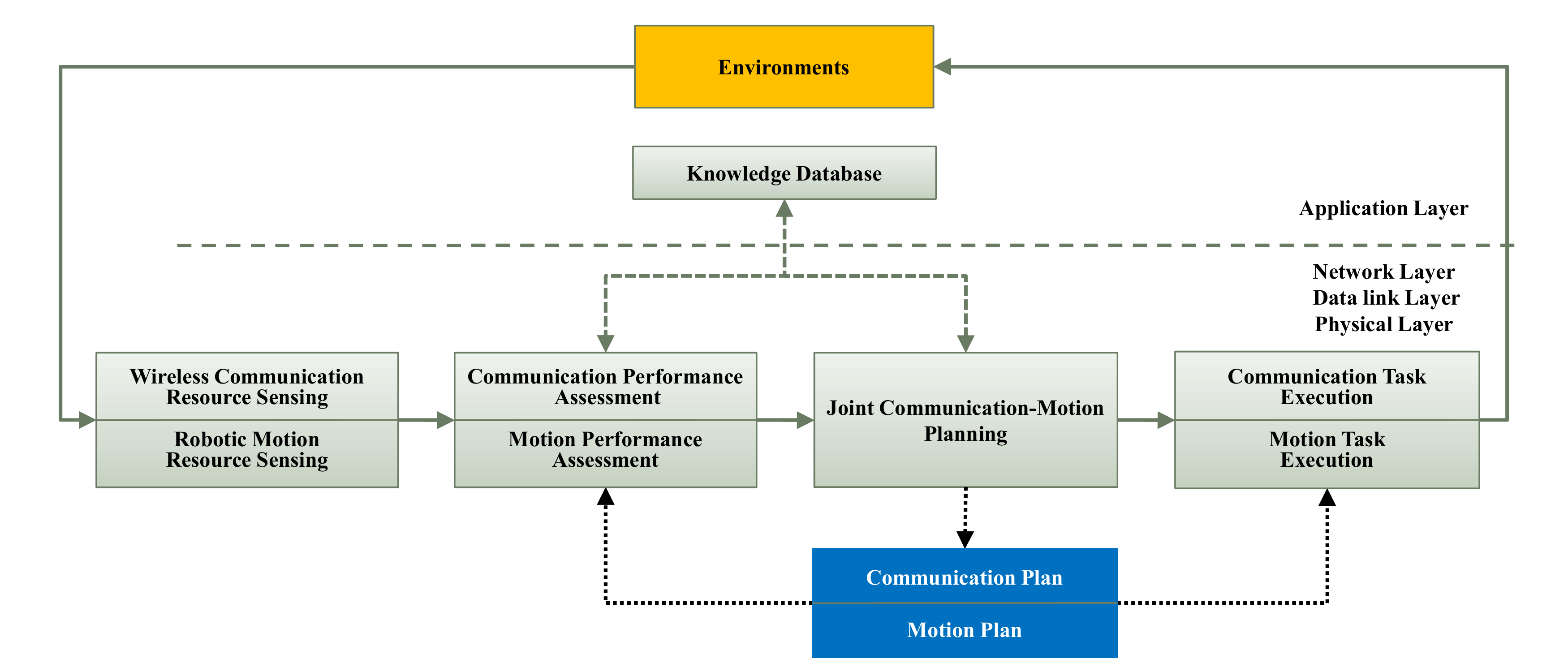}\protect\caption{The OODA-based architecture of the wireless-connected robotic networks
(WCRNs).}
\label{fig:OODA-WCRN}
\end{figure*}

Motivated by the OODA loops in the robotic motion control and the
cognitive radio, we propose the OODA architecture for WCRNs, which
is illustrated in in Fig. \ref{fig:OODA-WCRN}. Traditionally, the
two OODA loops are separately designed and implemented, while the
WCRN architecture in Fig. \ref{fig:OODA-WCRN} fuses both OODA loops
in motion control and cognitive radio by incorporating the JCMP, which
joint exploits the DoF in MSTFE dimensions. 

The information flow in Fig. \ref{fig:OODA-WCRN} is as follows. Firstly,
the wireless communication devices and the motion sensors implement
the communication and motion resource sensing, respectively. Then,
the measurements are imported to the respective performance assessment
modules and generate a performance quality metric according to the
mission objectives, given a communication/motion plan. Afterwards,
the JCMP module generates the optimized plan and request the task
execution module to take the corresponding actions. As a final remark,
the knowledge database supports the performance assessment and the
JCMP modules to store and learn from past experiences, as inherited
from CR. The information flow in Fig. \ref{fig:OODA-WCRN} indicates
that the connectivity quality assessment is a requisite for implementing
JCMP. Therefore, we provide a overview in the following section.

\section{Connectivity Quality Metric and Assessment}

\label{sec:connectivity_metric_assess}

Numerous research contributions have been devoted to quantifying the
quality of connectivity, which may be divided into two categories.
Firstly, the measure of connectivity over a graph is revisited and
its disadvantages are discussed in Section \ref{sub:CAMP:graph_theoretic_connectivity_control},
which motivates the research on the second category of realistic channel
model based connectivity quality metric (RCM-based CQM). For a more
detailed overview of graph-theoretic connectivity quality metric (GT-CQM),
please refer to \cite{Zavlanos2011}.

Compared to the family of graph-theoretic CQM, the family members
of RCM-based CQM are much more diverse, e.g. the bit error rate (BER),
packet error rate (PER), capacity, transmission rate, etc. These metrics
have been researched in the field of wireless communication for decades\cite{Goldsmith2005},
and they are task-specific and exhibit a fundamental trade-off, e.g.
diversity-multiplexing trade-off, etc \cite{El-Hajjar2010}. Therefore,
it is difficult and unnecessary to include a comprehensive review
over this category of connectivity quality metric. Instead, the SoAs
of applying RCM-based CQMs in WCRNs will be provided in Section \ref{sub:CAMP:realistic_channel_based_control}.

\subsection{Graph-theoretic CQM}

\label{sub:CAMP:graph_theoretic_connectivity_control}

As a robot may be treated as an intelligent agent, the communication
design for a WCRN is reflected in the research on connectivity in
multi-agent systems (MAS)\cite{Olfati-Saber2007}. The authors of
\cite{Ji2007,Zavlanos2007,Zavlanos2008,Zavlanos2011} have considered
the scenarios where a group of agents or robots are achieving some
mission objective, while addressing the problem of maintaining connectivity
during the mission. These contributions mainly adopts graph theory
for modeling the network, where the agents or robots are abstracted
into nodes in the graph, while the edges represent the communication
links between nodes\cite{Zavlanos2011}. Within this framework, the
most typical metrics for capturing the connectivity in the networks
are the algebraic connectivity metric and the number-of-path metric,
while both metrics have been widely adopted in a variety of scenarios,
e.g. exploration, surveillance, etc. For a comprehensive overview
and tutorial of adopting graph-theoretic definition of connectivity,
the readers are referred to \cite{Olfati-Saber2007,Zavlanos2011},
in which the authors also provided various approaches ranging from
convex optimization to potential fields based control methods in order
to optimize or maintain communication connectivity in MASs or WCRNs.

Specifically, a network of multiple communication links is modeled
by a weighted state-dependent graph, where each link between two nodes
$i$ and $j$ at time $t$ is associated with a weight commonly defined
as $w_{ij}\left(t\right)=f\left(\left\Vert x_{ij}\left(t\right)\right\Vert \right)$,
and $\left\Vert x_{ij}\left(t\right)\right\Vert $ is the euclidean
distance between the pair of nodes and the non-negative weight function
$f\left(\bullet\right)$ may be of arbitrary shape according to the
definition of connectivity. For example, a step-shape weight function
\begin{equation}
w_{ij}\left(t\right)=f\left(\left\Vert x_{ij}\left(t\right)\right\Vert \right)=\begin{cases}
1 & ,\;x_{ij}\left(t\right)<x_{th}\\
0 & ,\smash{\;x_{ij}\left(t\right)\geq x_{th}}
\end{cases}
\end{equation}
models a connectivity metric, which has perfect connection $w_{ij}\left(t\right)=1$
if the distance between two nodes $x_{ij}\left(t\right)$ is smaller
than a threshold value $x_{th}$, and lost connection completely otherwise.
The above function is very similar to the outage probability (OP)
generally adopted in the analysis and design of wireless networks\cite{Goldsmith2005},
where an outage or a connection failure occurs when the instantaneous
signal-to-noise ratio (SNR) is below a pre-defined value. However,
the graph-theoretic definition of connectivity may face the following
challenges in robotic networks with realistic wireless channels:
\begin{itemize}
\item \textcolor{black}{In general, the QoS over wireless channels cannot
be fully captured by a weight function of the distance between two
nodes. The distance only determines the path-loss, while the received
SNR is also characterized by multi-path fading effects\cite{Malmirchegini2012}. }
\item \textcolor{black}{It is difficult for the algebraic connectivity and
number-of-paths metrics to capture the end-to-end communication QoS
of links involving multiple hops and diversity in the space, time
and frequency domain\cite{Fink2013}. }
\item \textcolor{black}{Most current works rely on symmetric weights by
assuming a pair of communication links have identical communication
quality, which is not practical in asymmetric scenarios. For example,
in the unmanned arieal vehicle (UAV) systems, a highly asymmetric
data traffic is common, where a high sensing data rate and a low control
data rate may co-exist between the UAV and the remote control station\cite{Luo2015}. }
\end{itemize}
Therefore, although the graph-theoretic CQM has been successfully
applied in various applications, a class of CQM metrics which may
fully capture the performance over realistic wireless channels was
demanded.

\subsection{RCM-based CQM}

\label{sub:CAMP:realistic_channel_based_control}

Against the challenges exhibited in the graph-theoretic connectivity
control, the more realistic wireless channel models along with the
CQM for quantifying the communication quality-of-service (QoS) have
been introduced into the design and control of WCRNs since 2009. Differnet
CQMs may be adopted for physical, data-link and network layers, and
the CQMs may be roughly divided into categories that quantify reliability
and spectral efficiency, respectively.

In terms of reliability, the BER in uncoded schemes and PER in coded
schemes have been widely used CQM metric for quantifying transmission
reliability for a variety of physical layer protocols\cite{Goldsmith2005}.
BER was adopted as the CQM for WCRNs in \cite{Yan2012,Yan2013,Yan2014}
and PER was adopted in \cite{Tekdas2009,Ghaffarkhah2012}. In terms
of spectral efficiency, capacity and transmission rate are widely
used CQM and they were introduced to WCRNs in \cite{Lindhe2009,Ali2015}.
At higher protocol layers, the end-to-end (e2e) PER and e2e transmission
rate may reflect the performance\cite{Liu2004,Babaee2010}, which
were introduced to WCRN applications in \cite{Fink2012,Fink2013}.

It should be noted that although the definitions and application scenarios
of the above metioned reliability and spectral efficiency CQMs are
different, there is a fundamental trade-off so both categories of
CQMs may be interchangeable in terms of quantifying the connectivity
quality. For example, the diversity-multiplexing trade-off in wireless
system design from the information-theoretic perspective \cite{Tse2005a}
and the practical multi-functional multi-input multi-output system
design in order to strike the trade-off between spatial diversity,
multiplexing and beamforming\cite{Hanzo2011}.

\section{Joint Communication-Motion Planning}

\label{sec:JCMP}

In this section, we would investigate the third research question
proposed in Section \ref{sec:Introduction}. The JCMP design methods
are categorized into single- and multi-stage methods in Section \ref{sub:JCMP:overview},
followed by an illustrative example in Section \ref{sub:JCMP:example}.

\subsection{Single and Multi-Stage Methods}

\label{sub:JCMP:overview}

Let us revisit the OODA architecture in Fig. \ref{fig:OODA-WCRN}.
The WCRN implements both the communication and motion resource sensing,
where the observations are fed into the performance assessment module.
Then, given a specific communication/motion plan, the communication
and motion performance are evaluated and fed back to the JCMP module
so that an optimized or optimal plan may be found. 

From a mathematical perspective, the communication and motion performance
are quantified and formulated as \textit{\textcolor{black}{cost functions
}}(e.g. total energy consumption, etc.) relying on the chosen communication-motion
plans. The plans, on the other hand, is modeled as a set of control
\textit{\textcolor{black}{variables}}. Finally, the availability of
communication/motion resources (e.g. transmission power, time, bandwidth,
etc.) as well as the mission objectives (e.g. PER, video quality metric,
security, etc.) set several \textit{\textcolor{black}{constraints.}}
Therefore, for each OODA loop in WCRN, the JCMP module may formulate
an optimization problem in order to minimize the cost functions by
optimizing the variables, while satisfying certain constraints. 

The JCMP methods may be categorized into multi-stage and single-stage
methods. In multi-stage methods, the original problem is decomposed
into a collection of simpler sub-problems, where the dynamic programming
(DP) technique is well-known and widely adopted\cite{Chades2014}.
In comparison, various tools may be applied to solve the the single-stage
optimization problem, which may be seen as a special case of the multi-stage
counterparts having a single sub-problem. If the problem is convex,
the convex optimization tools may be the most efficient choice\cite{Hindi},
while a large family of heuristic optimization tools may be selected
for solving a non-convex problem\cite{Kulkarni2011}. 

In order to illustrate the differences between the single-stage and
multi-stage methods, a multi-robot surveillance example is provided
as follows.

\subsection{A Multi-Robot Surveillance Example}

\label{sub:JCMP:example}

We consider a simple WCRN scenario of two robots as illustrated in
Fig. \ref{fig:Example:System_Model}, where a sensing robot surveys
an area and transmits the collected data to a remote base station.
However, the distance spanning from the sensing robot to the base
station is long, so that the direct transmission cannot support the
required communication quality in terms of PER. Therefore, a router
robot is deployed for relaying the data transmission from the sensing
robot to the base station, where decode-and-forward (DF) is adopted\cite{Bletsas2007}.
In order to maintain the required sensing quality received at the
base station, the PER should be kept below a pre-defined threshold.
The objective is to minimize the total energy consumption of both
the sensing and the router robots. Because the sensing robot is assigned
to survey the area and track a target, the motion energy of the sensing
robot is assumed to be determined by the trajectory of the target
and cannot be optimized. Therefore, we focus on optimizing the sum
of the router robot's motion and communication energy as well as the
sensing robot's communication energy. 

The other simulation parameters are set as follows. We adopt the 802.11g
protocol as the communication specifications, where the corresponding
bandwidth is $B=20MHz$ and the noise power spectral density is $N_{0}=-100dBm/Hz$.
The pathloss exponent is $\beta=3.68$ and Rayleigh quasi-static fading
is assumed. Both the router and sensing robots are allowed to adapt
transmission power and the per-robot transmit power should be below
$4$ Watt. The transmission rate is also adaptive by selecting from
$6$ modes and the PER performance is modeled by the approximate expression
proposed by Liu et al\cite{Liu2004}. For the PER upper bound, we
use an accepted value $0.01$. The motion parameters are from the
Pioneer 3DX robot and $v_{max}=1m/s$. 

\begin{figure}[tb]
\begin{centering}
\includegraphics[width=1\linewidth]{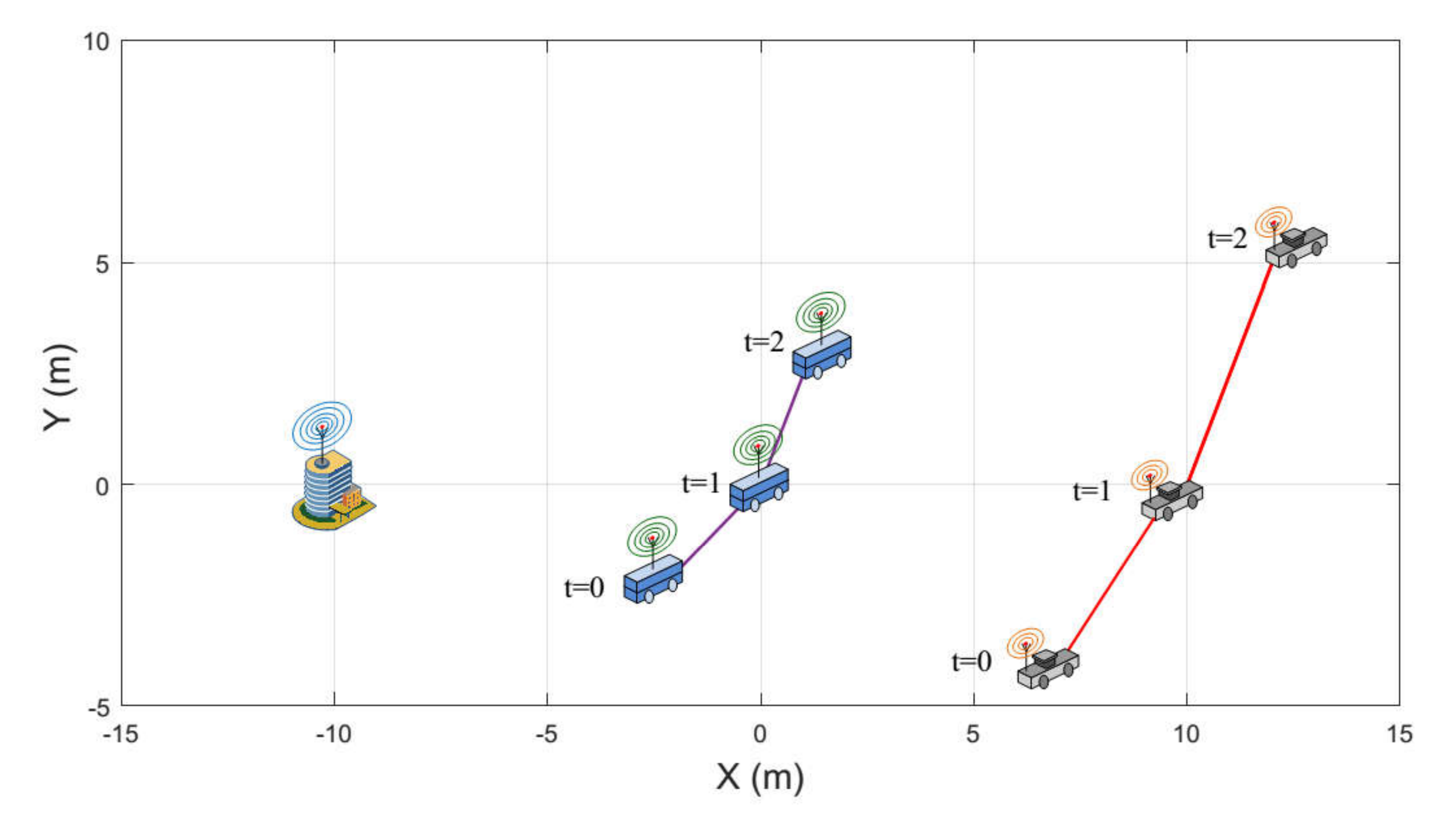}
\par\end{centering}

\begin{centering}
(a)
\par\end{centering}

\begin{centering}
\includegraphics[width=1\linewidth]{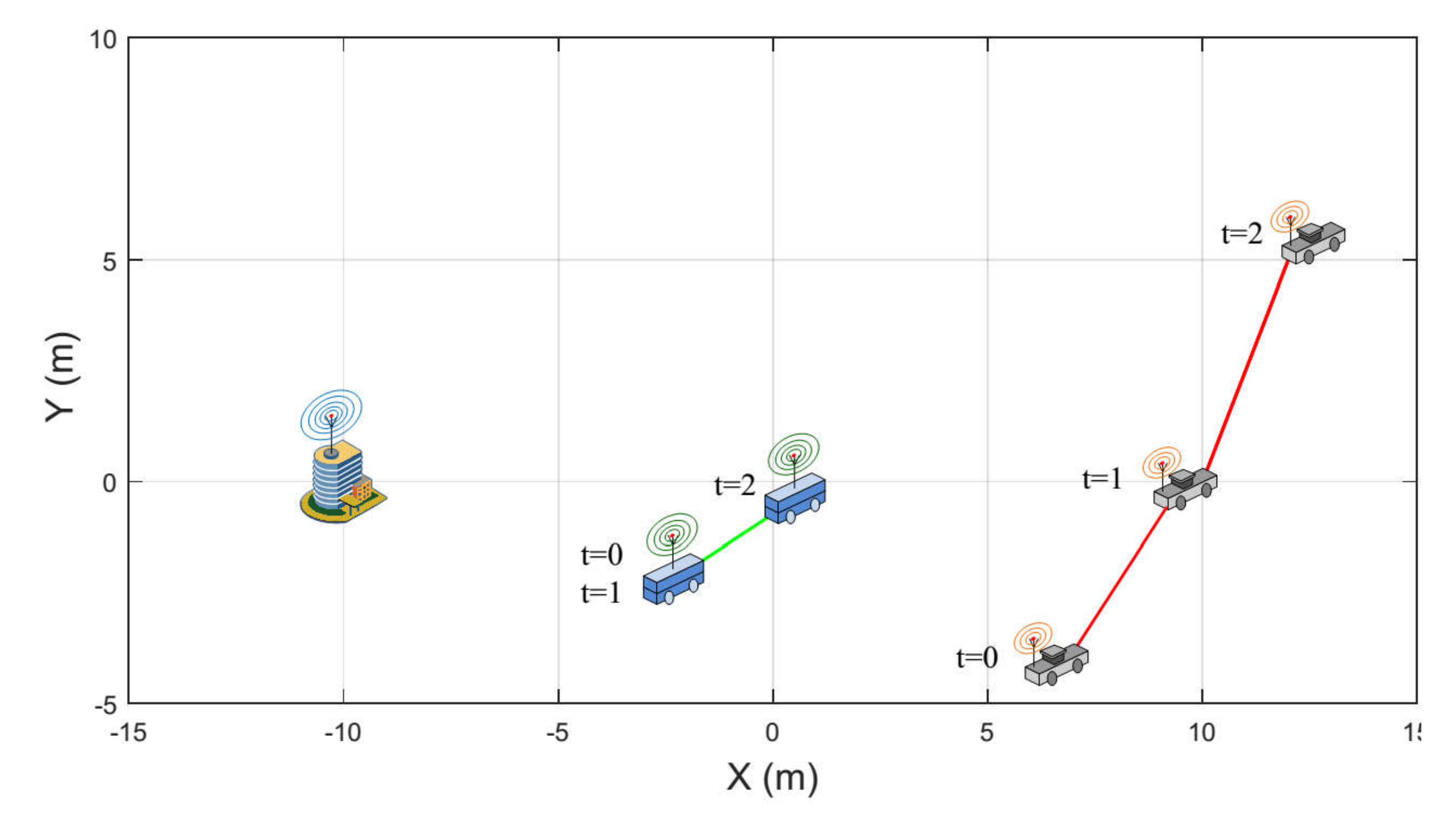}
\par\end{centering}

\begin{centering}
(b)
\par\end{centering}

\begin{centering}
\includegraphics[width=1\linewidth]{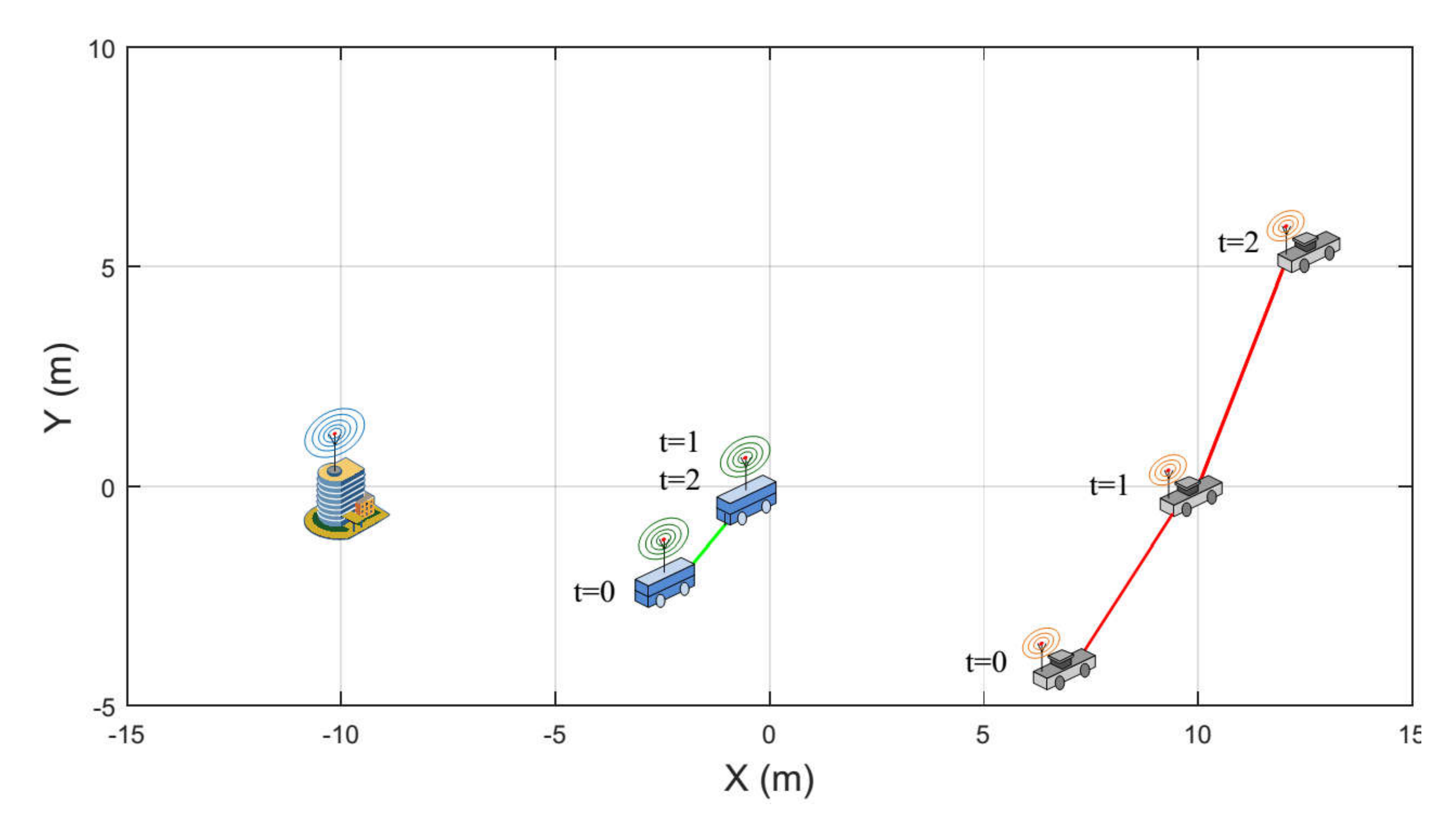}
\par\end{centering}

\begin{centering}
(c)
\par\end{centering}

\centering{}\protect\caption{From top to bottom: The comparison of the optimized trajectory of
the relay robot based on (b) communication planning (c) single-stage
JCMP (d) multi-stage JCMP by value iteration algorithm. }
\label{fig:Example:Static_Opt}
\end{figure}

The resulted trajectory of the router robot is given in Fig. \ref{fig:Example:Static_Opt}.
Fig. \ref{fig:Example:Static_Opt}(a) is the benchmark called communication-planning,
where no JCMP is implemented and the decide module in the OODA loop
only considers the minimization of communication energy, and the resulted
positions of the router robot is on the straight line spanning from
the sensing robot to the base station for each time step $t=0,1,2$,
which was also observed in \cite{Bletsas2007}.

Fig. \ref{fig:Example:Static_Opt}(b) shows the optimized trajectory
of the router robot with single-stage methods. Specifically, at the
beginning of each time step, the sensing robot is only capable of
predicting its own position in the current step according to the observations
of the unknown target. During each step, the JCMP may optimize the
router position in the current step, which is single-stage and the
plan consist of $3$ control variables, namely, the transmission power
of both robots and the position of the router robot. It is observed
that the router robot choose not to move in time step $t=1$. In this
way, the motion energy is conserved and the resulted total energy
saving for $2$ time steps is $16.9\%$ when compared to the benchmark. 

The bottom figure shows the optimized trajectory of the router robot
with multi-stage methods. Different from the previous cases, the sensing
robot is capable of predicting its own positions for the next two
steps. By assuming accurate prediction, the JCMP may optimize the
router position for two steps. As seen in Fig. \ref{fig:Example:Static_Opt}(c),
the router robots is planned to move in time step $t=1$ and keep
still in time step $t=2$. The different trajectory of the single-
and multi-stage methods is attributed to the availability of knowledge.
Compare to the single-stage component, the two-stage method achieves
a beneficial energy saving ratio of $17.7\%$ by exploiting the additional
knowledge. It should be noted that the energy saving comes at the
cost of a significant longer computation time, as in the multi-stage
problem, the dimension increases exponentially with the number of
stages and the solver are in general less computational-efficient
than their single-stage counterparts\cite{Powell2011}.

\section{Conclusions and Open Problems}

\label{sec:future_works}

In this technical note, we first reviewed the contributions on WCRNs,
with a focus on the CAMP scheme and proposed that JCMP may overcome
the disadvantages of CAMP. Then, we proposed the OODA-based WCRN architecture
that incorporates JCMP. After reviewing the SoAs in connectivity quality
metric for the performance assessment module, the JCMP was discussed
in more details and an illustrative example was provided to compare
the single- and multi-stage JCMP optimization methods. The final purpose
of this technical note is to discuss open problems, which need to
be addressed in order to fill the gap between the state-of-the-art
and a practical WCRN.
\begin{itemize}
\item Firstly, more research efforts need to be devoted to the CQM design
and assessment. Although various CQM have been proposed and applied
to WCRNs, the robotic applications requires task-oriented mission
performance metrics, e.g. sensing video delivery quality\cite{Ksentini2006,Huo2014},
detection error probability \cite{Ghaffarkhah2012} and networked-control
performance metric\cite{Zhang2013e}, etc. The other fundamental problem
is to provide accurate CQM assessment in practical environments. Most
contributions in the literature relies on statistical channel model
(including the example provided in Section \ref{sub:JCMP:example}),
which may over simplify the parameters affecting the CQM in practical
system and therefore may incur erroneous assessment that misleads
the JCMP design. In order to bridge the gap between theoretical CQM
analysis and practical CQM assessment, several contributions have
been proposed. For example, Halperin et al. proposed the PER/rate
assessment based on practical SNR measurements\cite{Halperin2010},
and Kudelski et al. designed a novel support-vector-machine (SVM)
based link quality estimation protocol\cite{Kudelski2014}. However,
most CQM methods proposed in the literature considered point-to-point
communications, while a WCRN involves multiple inter-connected robots.
Therefore, the CQM assessment in multi-access network having a more
sophisticated network topology remains an open problem.
\item Secondly, more advanced approaches should be applied in JCMP for practical
applications. Most publications adopted single-stage optimization
tools for solving the CAMP/JCMP problem as in \cite{Lindhe2009,Tekdas2009,Ghaffarkhah2011,Ghaffarkhah2012,Daniel2011,Kim2012,Fink2013,Kudelski2014,Gil2015}.
In order to utilize the past experience and the predicted knowledge,
multi-stage JCMP may be adopted and DP is a classic choice. However,
the computation complexity induced by the ``dimension curses in state
space and action space'' in DP may prohibit practical applications
for WCRNs, which incorporates multiple robots and exploits the DoF
from \textbf{\textit{MSTFE}} dimensions\cite{Powell2011}. Therefore,
the dimension reduction techniques from approximated dynamic programming
\cite{Powell2011} as well as from machine learning \cite{Bkassiny2013}
become indispensable. There have been some applications in WCRNs already
\cite{Bethke2009,Ure2015}, but the system models and the scenarios
considered are quite limited.  
\item Finally, the WCRN requires motion sensing, motion performance assessment
as well as motion execution as illustrated in Fig. \ref{fig:OODA-WCRN},
hence it demands a platform for implementation, testing and verification
in practical scenarios. The WCRN may be treated as one of the enabling
technologies in \textbf{\textit{collective robotics}}. The research
of collective robotics may involve interdisciplinary efforts in order
to deal with the technological, scientific, and social problems in
artificial and mixed societies consisting of many interacting entities,
which may be \textit{morphable} and \textit{intelligent}\cite{S.Ponda2015}.
Against this background, we proposed the morphable, intelligent and
intelligent robotic operating system (micROS)\cite{Yang2015}, which
is an open-source project and is available at http://micros.nudt.edu.cn.
The micROS is based on the robot operating system (ROS) project\cite{Quigley2009},
while focuses intensively on morphable resource management, autonomous
behavior management in order to support collective intelligence. The
micROS is also designed based on the OODA model and JCMP will be released
as one of the packages. In order to enable practical CQM assessment,
the other package under development supports soft-defined radio (SDR)
platform based on GNU-Radio\cite{Blossom2004}.
\end{itemize}
\vspace{-0cm}
\bibliographystyle{IEEEtran}
\bibliography{Energy_Aware_Bot}

\end{document}